\documentclass[letterpaper, 10 pt, conference]{ieeeconf}  

\IEEEoverridecommandlockouts                              
\usepackage{graphicx}
\usepackage{amsmath}
\usepackage{cite}
\usepackage{url}

\usepackage{times}
\usepackage{fancyhdr,graphicx,amsmath,amssymb}
\usepackage[ruled,vlined]{algorithm2e}
\include{pythonlisting}

\usepackage[font=footnotesize,belowskip=-0.5pt,aboveskip=0pt]{caption}
\usepackage{bigstrut}
\usepackage[utf8]{inputenc}
\usepackage{array}
\usepackage{multirow}
\usepackage{tabularx}
\usepackage{bigstrut}
\usepackage{makecell}
\usepackage{tabu}
\usepackage{bm}
\usepackage{caption}

\usepackage{amsmath}          
\usepackage{graphics}
\usepackage{graphicx}
\usepackage{float}
\usepackage{caption}
\usepackage{subcaption}
\usepackage{dblfloatfix}
\usepackage{calc}
\usepackage{url}
\usepackage{textcomp}
\usepackage{balance}
\usepackage{siunitx}
\usepackage{mdwlist}
\usepackage{comment}

\usepackage{gensymb}

\usepackage{array}
\usepackage{multirow}
\usepackage{tabularx}
\usepackage{booktabs}

\usepackage{bigstrut}
\usepackage{fixltx2e}

\usepackage{color}
\usepackage{hhline}

\usepackage[capitalize]{cleveref}

\title{\LARGE \bf
Strain in Sound: Soft Corrugated Tube for \\ Local Strain Sensing with Acoustic Resonance
}

\author{Michael Chun, Ananya Nukala, Tae Myung Huh
\thanks{All authors are with Dept. of Electrical and Computer Engineering, University of California Santa Cruz, Santa Cruz, CA, USA. {\tt\small thuh@ucsc.edu}}%
}

\begin{document}

\maketitle

\thispagestyle{empty}
\pagestyle{empty}

\global\csname @topnum\endcsname 0
\global\csname @botnum\endcsname 0

\begin{abstract}
We present a soft corrugated tube sensor designed to estimate strain in each half segment. When air flows through the tube, the internal corrugated cavities induce pressure oscillations that excite the tube’s standing wave resonance mode, generating an acoustic tone. Stretching the tube affects both the resonance mode frequency, due to changes in overall length, and the frequency-flow speed relationship, due to variations in cavity width, which is particularly useful for local strain estimation.
By sweeping flow rates in a controlled manner, we collected resonance frequency data across flow speeds under various local stretch conditions, enabling a machine learning algorithm (gradient boosting regressor) to estimate segmental strain with high accuracy. The dual-period tube design (3.1 mm and 4.18 mm corrugation periods) achieved a mean absolute error (MAE) of 0.8 mm, while the single-period tube (3.1 mm) provided a satisfactory MAE of 1 mm. Testing on a mannequin finger demonstrated the sensor’s capability to differentiate multi-joint configurations, showing its potential for estimating non-uniform deformations in soft bodies.

\end{abstract}



\section{Introduction}
\label{sec:Intro}

Soft robotic bodies, with their inherent nonlinear behavior, present challenges in estimating their deformed shape and pose. Although some methods for soft body modeling\cite{armanini2023soft}, control based on modeling\cite{della2023model}, and data-driven approaches such as the Koopman operator\cite{bruder2020data} have been explored, accurately describing shape becomes especially difficult when the body makes physical contact with the environment.

To address these challenges, various strain sensors for shape measurement have been developed for soft robotics, including stretchable electronics with resistive, capacitive, and optical sensors, as reviewed in \cite{souri2020wearable}. However, integrating these materials into soft robotics can be problematic, as they often differ fundamentally from the soft robotic material, leading to material mismatches and complex integration of electronic interconnects, which may result in fragile connections. Optical methods, often using embedded waveguides, also face challenges such as complex fabrication processes\cite{peng2019flexible}.

Most stretchable strain sensors in the literature capture only overall length changes in soft robotic bodies. However, when a robotic component contacts an object, creating localized curvature or strain in certain regions while leaving others unaffected, these sensors may generate misleading information, confusing small overall strain with large, concentrated strain. Embedding multiple strain sensors can provide localized strain information\cite{pinto2017cnt}, but it significantly increases system complexity.

Another transduction approach found in the soft robotics literature is acoustic sensing. Examples include contact-induced sound sensing with built-in microphones\cite{wall2023passive}, injecting sound and measuring acoustic wave dissipation for curvature sensing\cite{sofla2024soft}, and airflow generating acoustic resonance in a soft cavity to measure strain and touch\cite{li2024acoustac,li2022resonant}. This acoustic resonance method is intriguing as it allows for remote detection of resonant tone changes via an external microphone, simplifying data capture. However, the previous designs required additional structures like fipples and are limited to measuring overall length rather than segmented strains.

\begin{figure}[tbp!]
\centering
 
	\includegraphics[width=\linewidth]{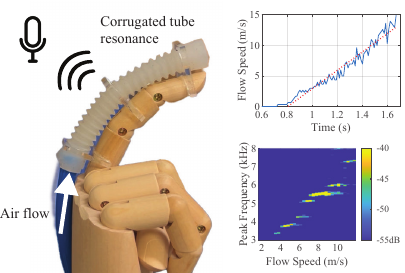}
    \vspace{+2pt}
	\caption{A soft corrugated tube generates acoustic resonance with airflow. By sweeping the flow speed (top right), the corresponding resonance frequency shifts (bottom right), which relates to the strain in each half segment.
}
	\label{fig:main}
	\vspace{-15pt}
\end{figure}

Inspired by acoustic resonance principles, we introduce a novel strain sensor utilizing a corrugated tube with airflow resonance. Corrugated tubes, characterized by wavy internal side walls, are commonly employed in flexible and bendable pipes. When air is blown through these tubes, they produce acoustic tones—famously used in toys like whirly tubes—and have been extensively studied in the context of airflow resonance \cite{rajavel2013acoustics}. This setup presents a promising solution for soft robotics: a highly stretchable and bendable tube made from the same material as the robot, capable of generating acoustic tones without additional structures like fipples. Additionally, localized strain on the corrugated tube alters the corrugation cavity sizes, which we found to be effective for detecting both segmented strain and overall length changes.

In the following sections, we review studies on the acoustic resonance of corrugated tubes and present the design of our soft corrugated tube sensor. We describe the experimental setup used to collect data on segmented stretch of the fabricated sensor and provide results including spectrogram analysis, peak resonance frequency variation with flow speed, and machine learning-based predictions of segmented stretch. Our sensor achieves independent estimation of the first and second half of its length with a mean absolute error (MAE) of less than 1 mm. Finally, as a demonstration, we test the sensor’s response to various finger joint configurations, as shown in \cref{fig:main}.

\section{Design}
\label{sec:Design}

\subsection{Corrugated Pipe Flow Acoustics}
\label{sec:theory}

The acoustic whistling response from airflow within corrugated pipes has been extensively studied in fluid dynamics; a comprehensive review is available in \cite{rajavel2013acoustics}. Prior research indicates that the resonant frequencies of whistling tones tend to increase with internal flow velocity, often 'locking in' to a harmonic mode of the pipe’s acoustic resonance.

\begin{figure}[tbp!]
\centering
	\includegraphics[width=2.4in]{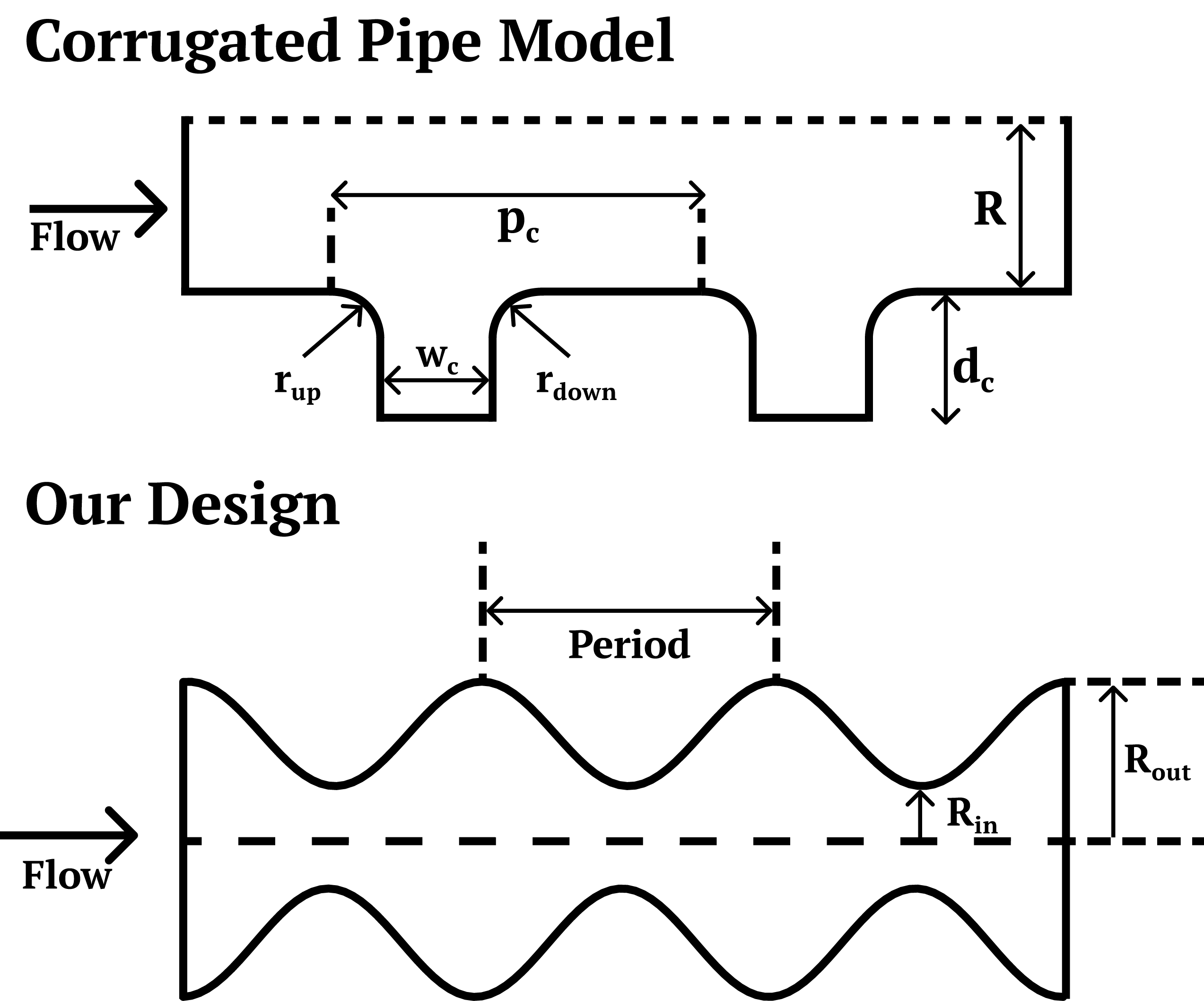} 
    \vspace{+6pt}
	\caption{(Top) Corrugated Pipe Model used in \cite{Wright2004LectureNO}. (Bottom) Our Design with cosine corrugation.}
	\label{fig:ModelvReal}
\end{figure}

The harmonic resonance of an open-ended pipe can be approximately estimated using the standing-wave relationship:
\begin{equation}
f_n = \frac{n c}{2L}
\label{eqn:baseResonance}
\end{equation}
where \( n \) is the mode number, \( c \) is the speed of sound, and \( L \) is the length of the tube. The impact of corrugated geometries can be considered using the Cummings Acoustic Model \cite{Wright2004LectureNO}:
\begin{equation}
f_n = \frac{n c}{2L} \left( \frac{1 - M^2}{1 + \frac{d_c}{R} \left( \frac{w_c}{p_c} \right) \left( 1 + \frac{d_c}{2R} \right)} \right)
\label{eqn:Resonance_corrugated}
\end{equation}
where \( M \) is the Mach number, \( d_c \) represents the cavity depth, \( R \) is the pipe radius, and \( w_c \) and \( p_c \) denote the cavity width and pitch, respectively (see \cref{fig:ModelvReal}). This model serves as a correction factor for geometry applied to the basic open-ended pipe resonance model in \cref{eqn:baseResonance}.
When the soft, corrugated pipe stretches, the primary effect on the resonance mode arises from the increase in \( L \), which lowers the fundamental mode of resonance. Additionally, changes in the cavity width and depth due to stretching influence the resonant frequencies.

\begin{figure}[tbp!]
\centering
	\includegraphics[width=0.75\linewidth]{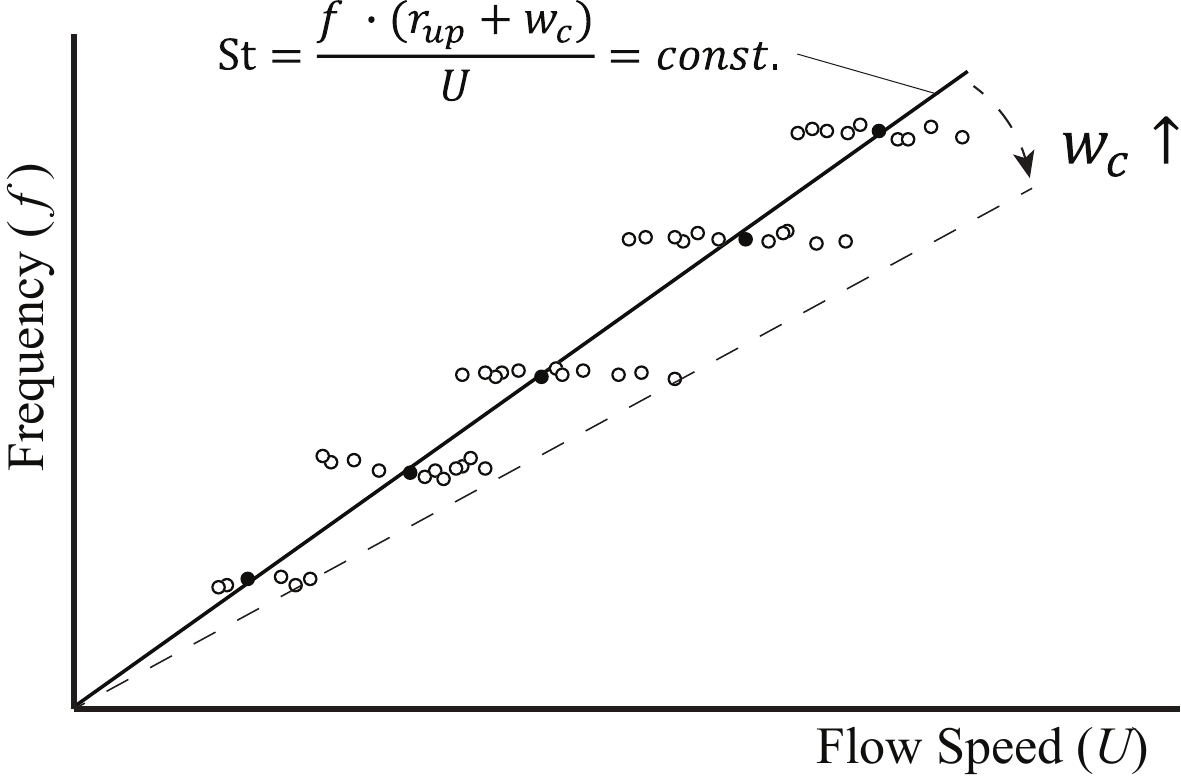}
    \vspace{+2pt}
	\caption{Conceptual acoustic resonance plot adapted from \cite{nakamura1991sound,nakibouglu2010whistling}. The acoustic tones lock into (plateau at) the resonance modes, producing strong resonances (solid dots) where the linear Strouhal number line intersects the resonance modes. }
	\label{fig:concept_literature}
	\vspace{-15pt}
\end{figure}

Among the resonance modes, the excitation of a particular mode is determined by the flow speed. This corrugated pipe whistles without any fipple structure, as the main pipe flow and internal cavity induce an impinging shear-layer instability, where vortex shedding oscillates the pressure \cite{rajavel2013acoustics, nakamura1991sound}. The frequency of this vortex shedding is linearly proportional to the flow speed, and when it matches the tube’s resonance mode frequency, it generates an audible acoustic tone \cite{nakamura1991sound}, as illustrated in the conceptual plot in \cref{fig:concept_literature}. The plateau region in the figure indicates that the audio frequencies are locked into the resonance mode frequency, allowing the vortex shedding frequency to excite adjacent resonance modes.

The linear relationship between flow speed (\( U \)) and the vortex shedding frequency (\( f_v \)) can be understood using the Strouhal number, defined as
\begin{equation}
\text{St} = \frac{f_v \cdot L}{U}
\label{eqn:StrouhalNum}
\end{equation}
where \( L \) represents a characteristic length. Research has shown that when \( L \) is taken as the sum of cavity width (\( w_c \)) and edge roundness (\( r_up \)), the Strouhal number remains relatively constant across a variety of cavity pitches and widths\cite{nakibouglu2010whistling}, indicating that the slope of \( f_v \) versus \( U \) is predominantly determined by the cavity opening width. In summary, when the soft, corrugated tube is stretched, resulting in an increase in both the overall length (\( L_{\text{all}} \)) and cavity width (\( w_c \)), the resonance frequency is expected to decrease, and the \( f_v \) versus \( U \) slope is expected to decrease as well.

\vspace{+6pt}

\subsection{Design and Fabrication}
\begin{figure}[tbp!]
\centering
	\includegraphics[width=2.2in]{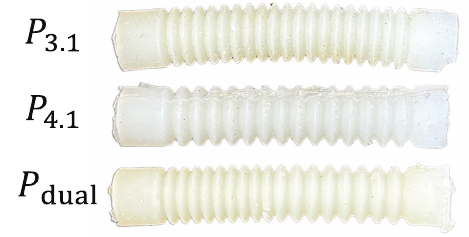}
    \vspace{+2pt}
	\caption{Three fabricated sensors with different periods. \( P_{3.1} \) has a period of 3.14mm, \( P_{4.1} \) has a period of 4.18mm, and \( P_{\text{dual}} \) has both the 3.14 mm (left) and 4.18 mm (right) periods split at the sensor’s midpoint.}
	\label{fig:SensorFabs}
	\vspace{-10pt}
\end{figure}

Given that the rounded edge of a cavity generates greater acoustic power \cite{nakibouglu2010whistling}, we designed the corrugated tube with a cosine wave pattern, as shown in \cref{fig:ModelvReal}, aligning the effective cavity width and the period of corrugation closely. The corrugation for each sensor was modeled using the following equation, where \( k \in \mathbb{R}_{\ge 0} \):
\begin{equation}
y = \cos\left(kx \right)
\end{equation}
We fabricated three sensors, shown in \cref{fig:SensorFabs}, with periods of 3.14 mm (\( k = 2 \)), 4.18 mm (\( k = 1.5 \)), and a dual-period design featuring 3.14 mm and 4.18 mm periods split at the sensor’s midpoint, denoted as \( P_{3.1} \), \( P_{4.1} \), and \( P_{\text{dual}} \), respectively. The inner and outer radii were kept consistent across all three sensors, with \( R_{\text{out}} = 3.9 \) mm and \( R_{\text{in}} = 1.9 \) mm. 

\begin{figure}[tbp!]
    \centering 
    
    \vspace{2mm} 
    
    \begin{subfigure}[b]{0.49\linewidth}
        \centering
        \includegraphics[width=0.8\linewidth]{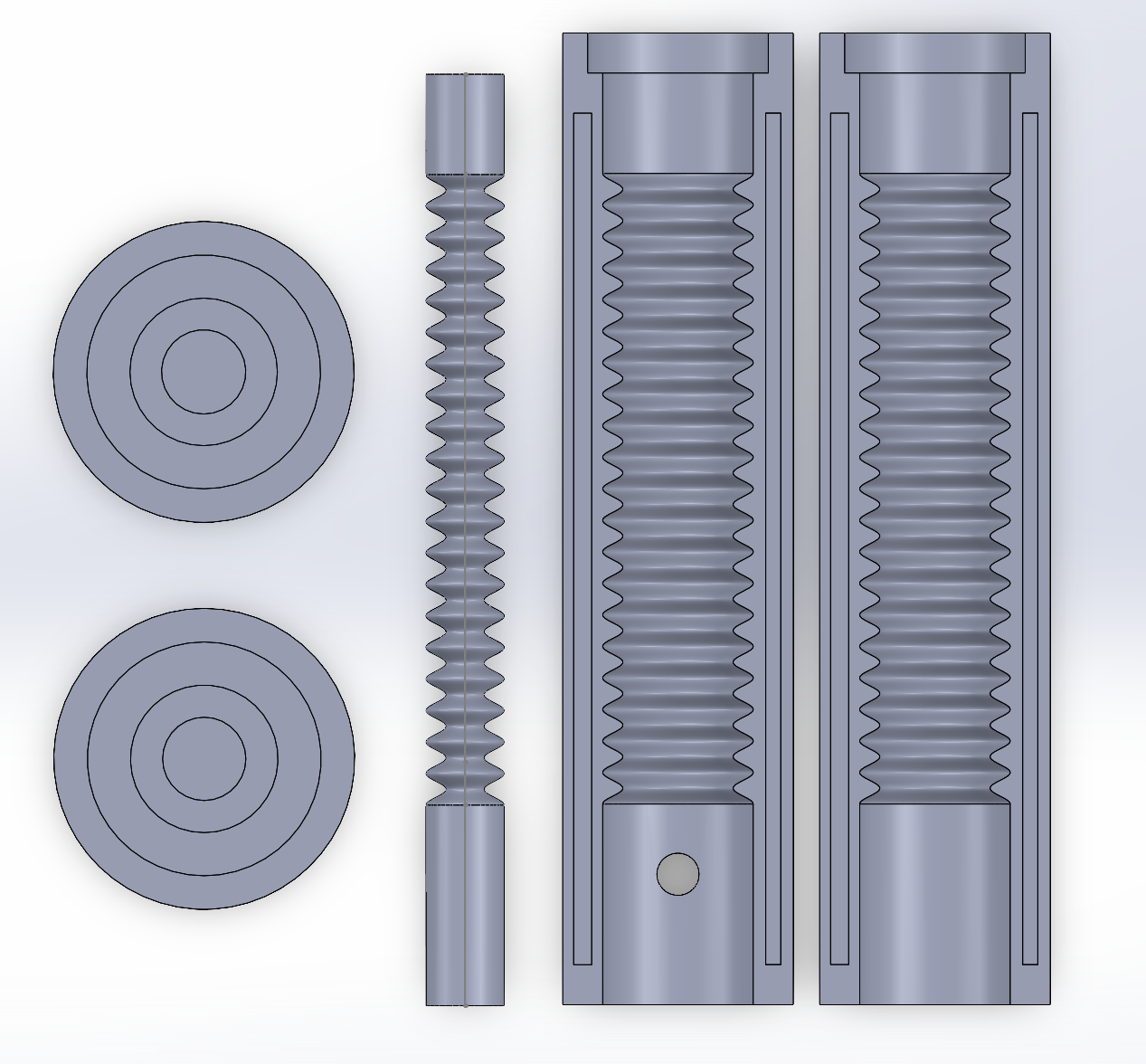}
        \caption{Five Part Mold}
        \label{fig:ComponentLayout}
    \end{subfigure}%
    \hspace{-2pt} 
    \begin{subfigure}[b]{0.20\linewidth}
        \centering
        \includegraphics[width=0.9\linewidth]{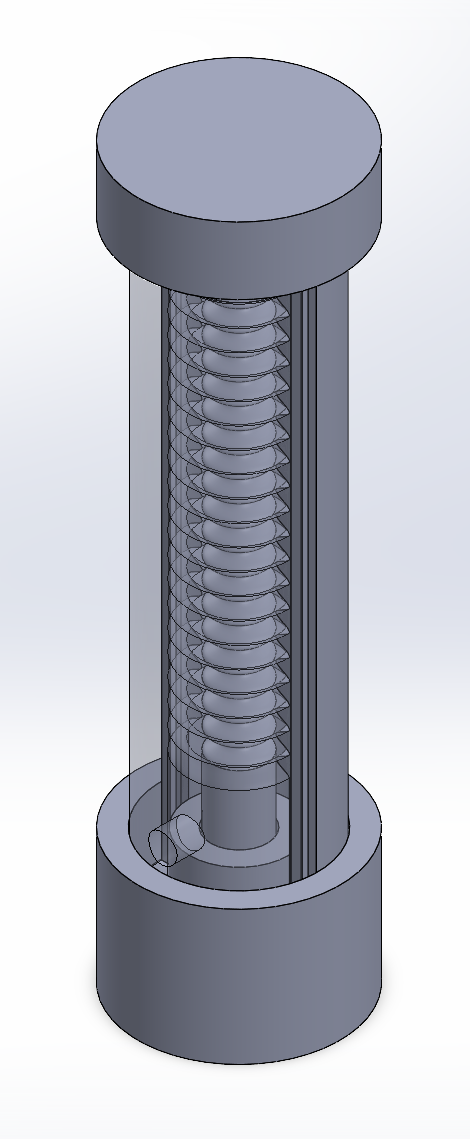}
        \caption{Inside View}
        \label{fig:AssemblyIsoClear}
    \end{subfigure}%

    \begin{subfigure}[b]{0.5\linewidth}
        \centering
        \includegraphics[width=0.8\linewidth]{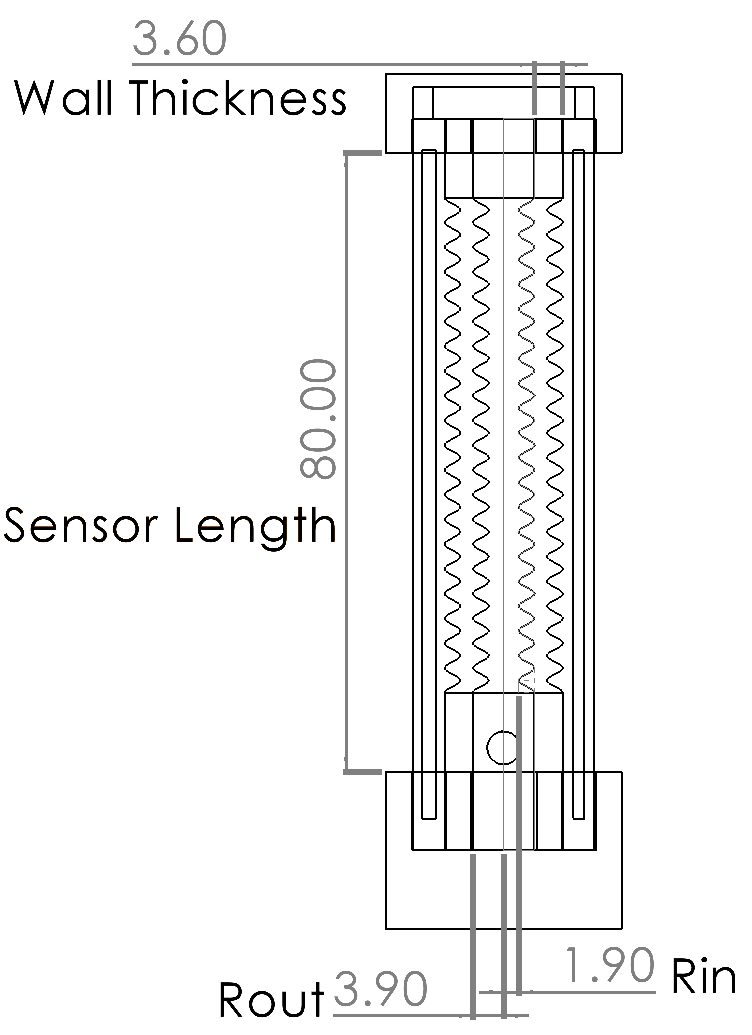}
        \caption{Mold Dimensions (mm)}
        \label{fig:AssemblyDrawing}
    \end{subfigure}%
    \hspace{-2pt} 
    \begin{subfigure}[b]{0.25\linewidth}
        \centering
        \includegraphics[width=0.8\linewidth]{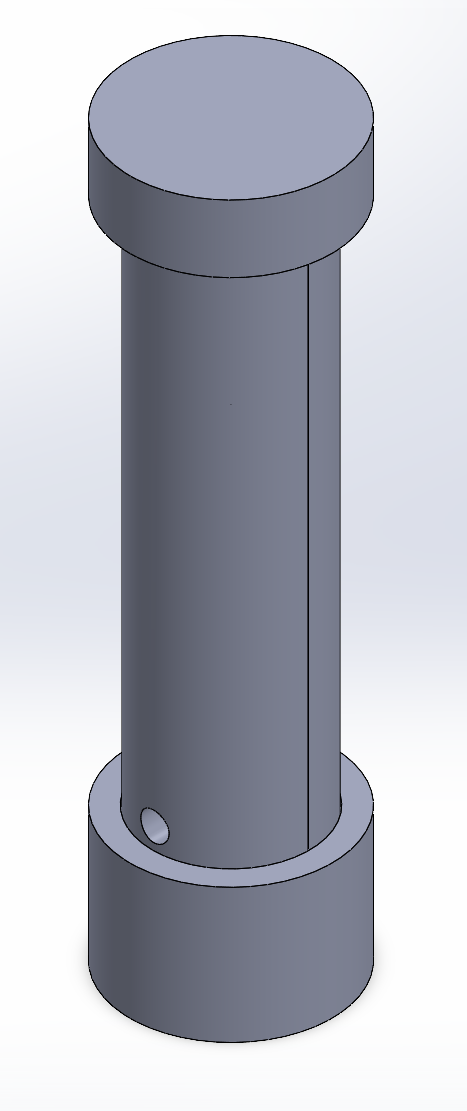}
        \caption{Assembled Mold}
        \label{fig:AssemblyIso}
    \end{subfigure}
    
    \vspace{5pt} 
    \caption{Overview of mold components and assembly.}
    \label{fig:SensorMold}
    \vspace{-2pt}
\end{figure}

Each sensor was made from Ecoflex 00-30 platinum cure silicone. To cast the silicone, we designed a five-part mold, depicted in \cref{fig:ComponentLayout}, which was 3D printed in polylactic acid (PLA). Each mold included the corrugated section and two flat sections at the widest radius (\( R_{\text{out}} \)). After assembling the mold, we injected uncured silicone through a bottom inlet hole using a syringe to minimize air bubbles. Finally, the mold was placed in a vacuum chamber for further degassing before removing the fully cured silicone.


\captionsetup{justification=centerlast} 

\captionsetup{justification=justified}


\section{Experiment}
\subsection{Test setup}

\begin{figure}[tbp!]
\centering
	\includegraphics[width=\linewidth]{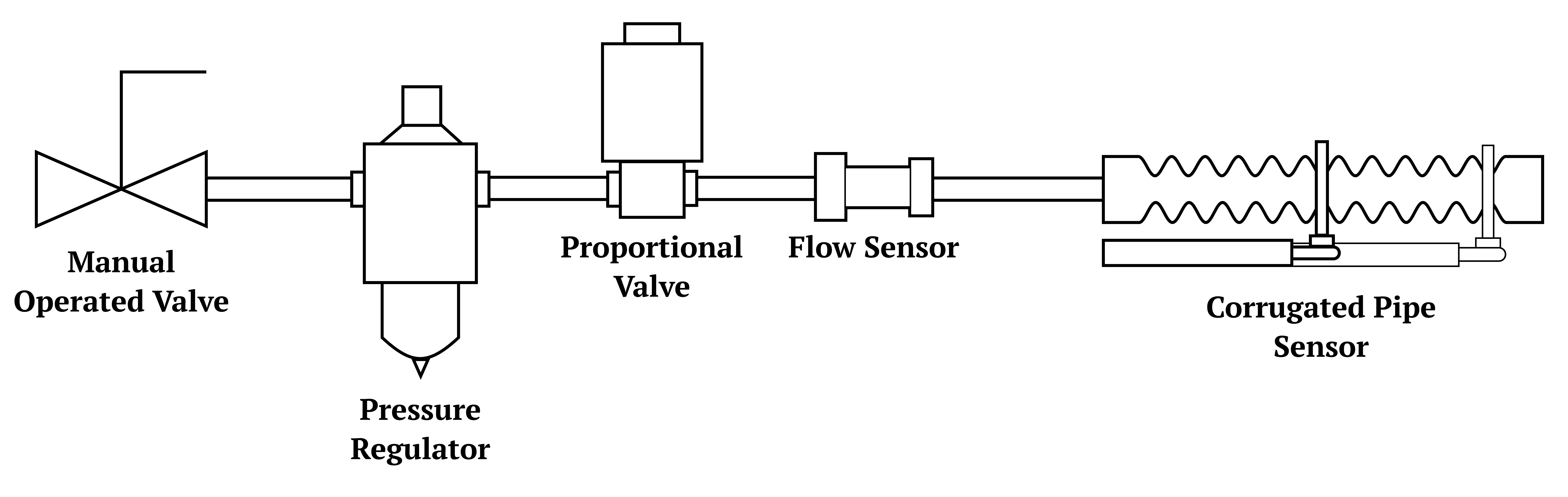}
	\caption{Data collection test setup. The air comes from a manually operated laboratory lab supply. This connects to a pressure regulator that sets the input pressure to 350kPa when the outlet is closed. We use a proportional valve to control flow rate, while we measure flow rate with a flow meter.}
	\label{fig:TestHarness}
\end{figure}

\begin{figure}[tbp!]
\centering
	\includegraphics[width=0.9\linewidth]{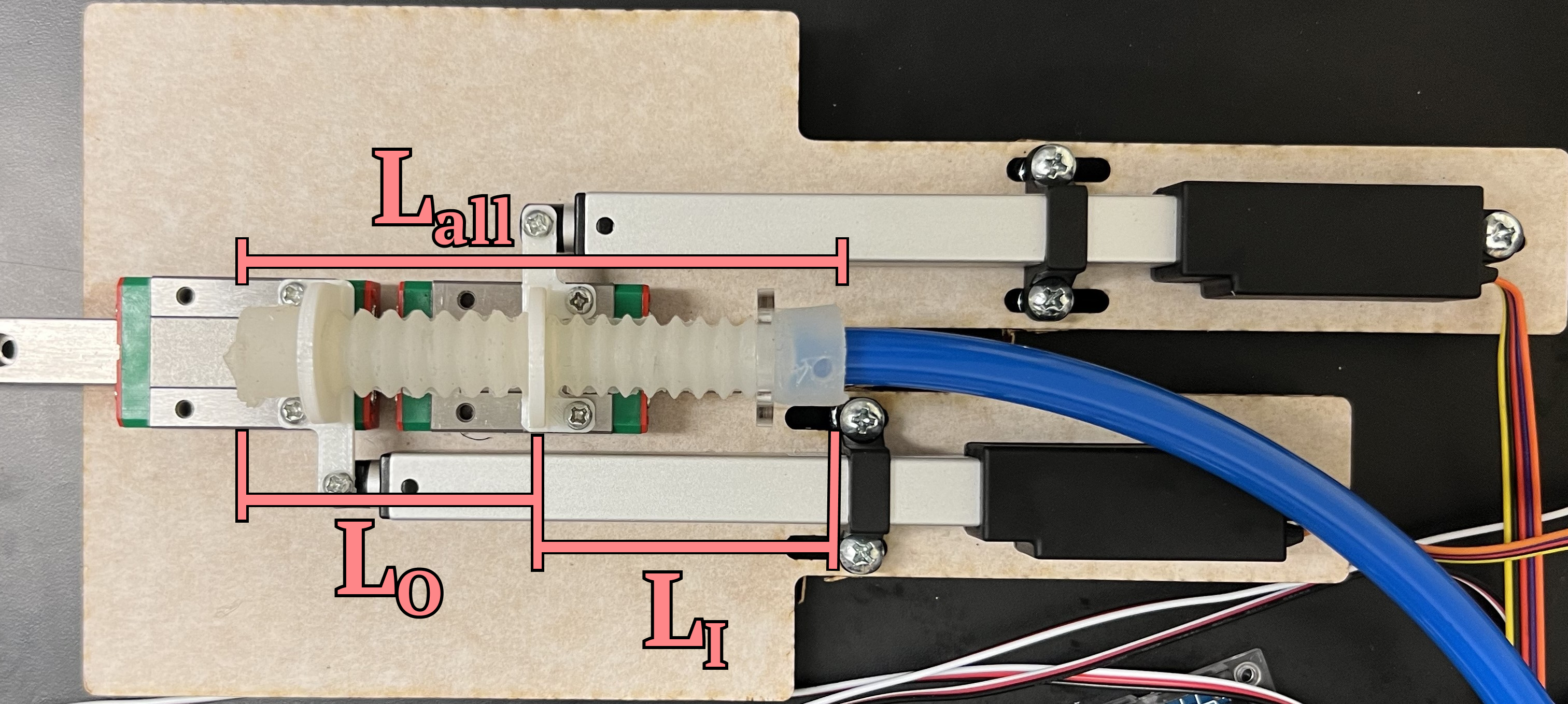}
    \vspace{+2pt}
	\caption{Sensor stretch harness to independently change lengths \(L_I\) and \(L_O\) where \(L_{all} = L_I + L_O\). Lengths change with linear actuators guided by a linear-rail.}
	\label{fig:TestHarness_TopView}
    \vspace{-5pt}
\end{figure}

We tested the acoustic resonances of the three fabricated sensors by varying the flow speed through the corrugated tube. To control and measure the flow rate, we built a flow test setup as shown in \cref{fig:TestHarness}. The setup utilizes compressed air from the laboratory wall supply, with a pressure regulator setting the input pressure to 350 kPa when the outlet is closed. The flow rate is controlled by a proportional valve (iQ Tesla Proportional Valve, iQvalves), operated via a valve controller board (iQ Signal Amplifier Valve Driver, iQvalves) connected to a microcontroller’s digital-to-analog converter (DAC).
In each test, the flow rate was gradually increased from fully closed to fully open over a 1.2-second period, with an effective change period of 0.8 seconds after the valve headband, as shown in \cref{fig:main}. During the flow ramp-up, we measured the actual flow rate every 10 ms using a flow meter (SFM3300-D, Sensirion).
To apply the desired strain to each half segment of the tube sensor, we constructed a linear-rail guided station, shown in \cref{fig:TestHarness_TopView}, with two independently controlled linear actuators (L16-P, Actuonix), allowing precise strain adjustment for each segment, $L_I$ and $L_O$. After setting the strain on each segment, we began audio recording on a laptop, activated a momentary buzzer for time synchronization, and then initiated the flow rate ramp-up.
In the case \( P_{\text{dual}} \), the direction of flow does affect resonance. Every experiment for \( P_{\text{dual}} \), was done with the flow coming from the side with a period of 3.14mm.

\subsection{Data Collection of Uniform and Segmented Strain Tests}

For each sensor (\( P_{3.1} \), \( P_{4.1} \), and \( P_{\text{dual}} \)), we collected data by stretching each segment, (\( \Delta L_I \) and \( \Delta L_O \)), within the range of [0, 10] mm. Beyond this range, the acoustic resonance was not detectable.
In the first test, we systematically varied each segment’s stretch (\( \Delta L_I \) and \( \Delta L_O \)) in 1 mm increments across the test range, resulting in 121 data points. Each measurement was repeated twice. In the second test, we randomly selected values for \( \Delta L_I \) and \( \Delta L_O \) within the test range to generate 100 additional data points, also repeated twice. This yielded a total of 442 recordings per sensor.

\subsection{Application for Finger Joint Angle Differentiation}

We applied our corrugated sensor in a data glove application, where the strain sensor measures each joint angle to estimate finger motion—a task that typically requires multiple strain sensors per joint \cite{jha2020design}. To replicate these conditions, we attached the sensor to a mannequin hand, as shown in \cref{fig:main}, and simulated joint movement by bending the joints to stretch the sensor. Using the \( P_{3.1} \) sensor, we tested three distinct finger poses: bending both joints, bending only the proximal joint, and bending only the distal joint.




\section{Result}

\subsection{Testing Resonance of three corrugated sensors}
\label{sec:baseResonance_neutral}

\begin{figure}[tbp!]
    \centering 

    	\begin{subfigure}[h]{\linewidth}
	\centering
	\includegraphics[width=0.75\linewidth]{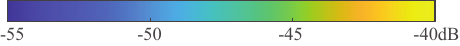}	
        \end{subfigure}
    \vspace{-2mm}

    \begin{subfigure}[h]{0.49\linewidth}
        \centering
        \includegraphics[width=0.9\linewidth]{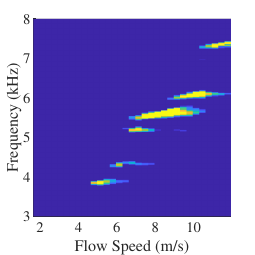}
        \caption{\( P_{3.1} \)}
    \end{subfigure}%
    \begin{subfigure}[h]{0.49\linewidth}
        \centering
        \includegraphics[width=0.9\linewidth]{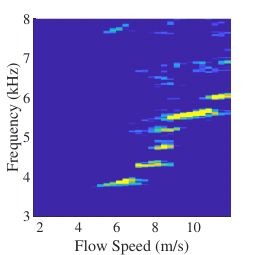}
        \caption{\( P_{4.1} \)}
    \end{subfigure}
    
    \vspace{5pt} 
    \begin{subfigure}[h]{0.49\linewidth}
        \centering
        \includegraphics[width=0.9\linewidth]{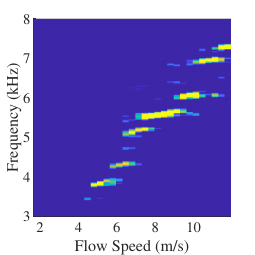}
        \caption{\( P_{dual} \)}
    \end{subfigure}%
    \begin{subfigure}[h]{0.49\linewidth}
        \centering
        \includegraphics[width=0.9\linewidth]{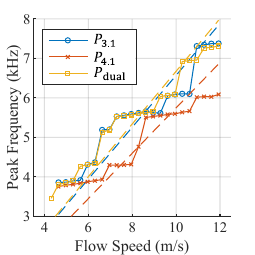}
        \caption{Peak Resonance Frequency}
    \end{subfigure}

    \vspace{5pt} 
    \caption{(a-c) Spectrograms of acoustic resonance at the neutral length of each sensor, plotted against flow speed. (d) Peak resonance frequency at each sampled flow speed, with a dashed line representing a zero-intercept linear fit. The slopes are 0.66, 0.57, and 0.67 kHz/(m/s), corresponding to the legend order.
    }
    \vspace{-2mm}
    \label{fig:threeSensor_nominalValue}
\end{figure}

We collected data on acoustic resonance as a function of inlet flow speed. \cref{fig:threeSensor_nominalValue}a–c show spectrograms with locked-in resonance frequencies that increase with flow speed. The fundamental resonance mode calculated from \cref{eqn:Resonance_corrugated} is 722 Hz for all three sensor configurations. Assuming identical cavity openings (\( w_c \)) and pitch (\( p_c \)) in our sensor design, the corrugation pitch effect cancels out in \cref{eqn:Resonance_corrugated}, resulting in identical resonance frequencies for all sensors. As flow speed rapidly varied, lower frequency ranges did not resonate strongly, but frequencies between 3 kHz and 8 kHz generated noticeable tones. The locked-in, or plateau, frequencies were similar across sensors, with occasional unique frequencies observed for each sensor. A detailed computational fluid dynamics (CFD) analysis may be required in future work to fully investigate geometry-specific vortex shedding and resonance.

The peak resonance frequency at each flow speed (\cref{fig:threeSensor_nominalValue}d) shows that the shorter pitch sensor (\( P_{3.1} \)) has a steeper frequency–flow speed (\( F\text{-}U \)) slope compared to the longer pitch (\( P_{4.1} \)) sensor, which aligns with expectations based on prior work, as discussed in \cref{sec:theory}. Interestingly, the dual-pitch sensor (\( P_{\text{dual}} \)) exhibits a similar \( F\text{-}U \) slope to the \( P_{3.1} \) sensor. We hypothesize that this occurs because vortices formed in the inlet region’s shorter-pitch corrugations have a smaller size that matches the cavity width \cite{nakibouglu2010whistling} and can survive in the downstream wider cavities. This interpretation is supported by the observation that reversing the inlet and outlet of the \( P_{\text{dual}} \) sensor eliminated the noticeable acoustic tone, likely because the larger vortices generated in the wider cavity do not survive or fit in the narrower downstream cavity. However, a CFD analysis is needed to substantiate this further in future work.

\begin{figure}[tbp!]
    \centering 

    \begin{subfigure}[h]{\linewidth}
	\centering
	\includegraphics[width=0.70\linewidth]{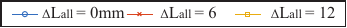}	
        \end{subfigure}
    \vspace{-2mm}

    \begin{subfigure}[h]{\linewidth}
	\centering
	\includegraphics[width=\linewidth]{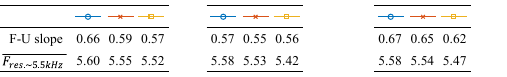}	
        \end{subfigure}
    \vspace{-0.5mm}
        
    \begin{subfigure}[h]{0.32\linewidth}
        \centering
        \includegraphics[width=\linewidth]{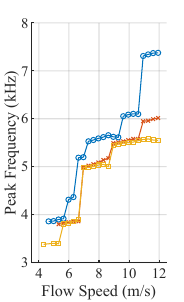}
        \caption{\( P_{3.1} \)}
    \end{subfigure}%
    \begin{subfigure}[h]{0.32\linewidth}
        \centering
        \includegraphics[width=\linewidth]{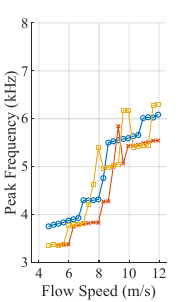}
        \caption{\( P_{4.1} \)}
    \end{subfigure}%
    \begin{subfigure}[h]{0.32\linewidth}
        \centering
        \includegraphics[width=\linewidth]{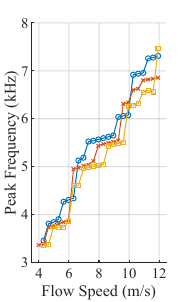}
        \caption{\( P_{\text{dual}} \)}
    \end{subfigure}%

    \vspace{5pt} 
    \caption{Uniform stretch test results showing peak resonance frequency at each flow speed.  (Top) Table shows the frequency-flow speed (\( F\text{-}U \)) slope in kHz/(m/s) and the mean resonance frequency around 5.5 kHz.}
    \label{fig:uniformStretch}
    \vspace{-10pt} 
\end{figure}

\begin{figure}[tbp!]
    \centering 
    \begin{subfigure}[h]{\linewidth}
	\centering
	\includegraphics[width=0.85\linewidth]{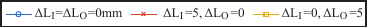}	
        \end{subfigure}
    \vspace{-2mm}

    \begin{subfigure}[h]{\linewidth}
	\centering
	\includegraphics[width=\linewidth]{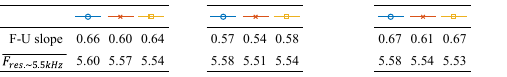}	
        \end{subfigure}

    \begin{subfigure}[h]{0.32\linewidth}
        \centering
        \includegraphics[width=\linewidth]{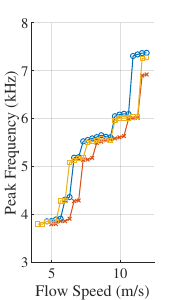}
        \caption{\( P_{3.1} \)}
    \end{subfigure}%
    \begin{subfigure}[h]{0.32\linewidth}
        \centering
        \includegraphics[width=\linewidth]{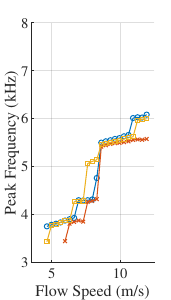}
        \caption{\( P_{4.1} \)}
    \end{subfigure}%
    \begin{subfigure}[h]{0.32\linewidth}
        \centering
        \includegraphics[width=\linewidth]{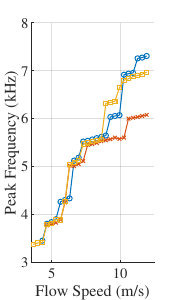}
        \caption{\( P_{\text{dual}} \)}
    \end{subfigure}%

    \vspace{5pt} 
    \caption{Segmented stretch test results showing peak resonance frequency at each flow speed. (Top) Table shows the frequency-flow speed (\( F\text{-}U \)) slope in kHz/(m/s) and the mean resonance frequency around 5.5 kHz.}
    \label{fig:segmentedStretch}
\end{figure}

\begin{figure}[tbp!]
    \centering 
    \begin{subfigure}[h]{\linewidth}
	\centering
	\includegraphics[width=0.6\linewidth]{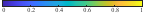}	
        \end{subfigure}
    \vspace{-3mm}

    \begin{subfigure}[h]{0.32\linewidth}
        \centering
        \includegraphics[width=\linewidth]{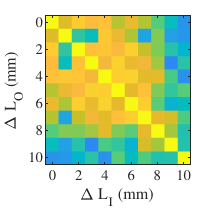}
        \caption{\( P_{3.1} \)}
    \end{subfigure}%
    \begin{subfigure}[h]{0.32\linewidth}
        \centering
        \includegraphics[width=\linewidth]{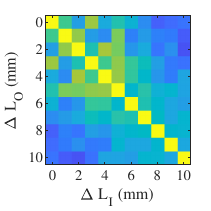}
        \caption{\( P_{4.1} \)}
    \end{subfigure}%
    \begin{subfigure}[h]{0.32\linewidth}
        \centering
        \includegraphics[width=\linewidth]{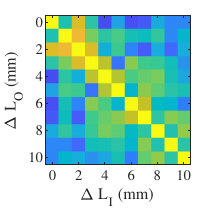}
        \caption{\( P_{\text{dual}} \)}
    \end{subfigure}%

    \vspace{5pt} 
    \caption{Similarity between spectrograms from mutual segmented stretch pairs, e.g., (\( \Delta L_O \), \( \Delta L_I \)) = (2, 5) mm and (5, 2) mm, resulting in symmetry across the diagonal. Mean correlation values for \( P_{3.1} \), \( P_{4.1} \), and \( P_{\text{dual}} \) are 0.70, 0.47, and 0.54, respectively. The lower similarity for \( P_{4.1} \) is likely due to reduced resonance, as reflected in its higher statistical entropy values, with average entropies of 8.8, 9.3, and 8.5, respectively.}

    \label{fig:similarity}
\end{figure}

\subsection{Uniform and Segmented Stretch Testing}
\subsubsection{Uniform Stretch}
\cref{fig:uniformStretch} illustrates that, under uniform stretch, the resonance frequency (e.g., approximately 5.5 kHz) of each of the three sensors decreases, and the frequency–flow speed (\( F\text{-}U \)) slope also reduces, consistent with the theory in \cref{sec:theory}. The increase in overall length lowers the tube resonance frequencies, as predicted by \cref{eqn:Resonance_corrugated}, while the uniform stretch increases cavity width, leading to a lower \( F\text{-}U \) slope according to \cref{eqn:StrouhalNum}. In the \( P_{4.1} \) sensor, more extensive stretching results in weak resonance, which we attribute to the stretched cavity failing to generate vortex shedding.

\subsubsection{Segmented Stretch}
\label{sec:segmentedStretch}
\cref{fig:segmentedStretch} compares two cases where the overall stretch length is the same, but only the inlet-side or outlet-side half segment is stretched. When assessing the effect of segment-specific stretching, changes in resonance frequency are not reliable indicators, as they are mostly affected by the total tube length. Instead, the frequency–flow speed (\( F\text{-}U \)) slope provides a useful feature that differentiates between inlet and outlet strain. When the inlet side is stretched, the \( F\text{-}U \) slope decreases as expected. However, when only the outlet side is stretched, with no inlet side stretch, the \( F\text{-}U \) slope remains largely unaffected, supporting our hypothesis in \cref{sec:baseResonance_neutral} that inlet side vortex shedding is the dominant contributor. This distinct behavior in the \( F\text{-}U \) slope under segmented stretch provides valuable information for estimating the stretch in each segment, alongside changes in resonance frequency.

To determine which sensor design is better suited for segmented length estimation, we analyzed the similarity between spectrograms from mutual segmented stretches by calculating their correlation. As shown in \cref{fig:similarity}, the \( P_{3.1} \) sensor exhibits higher similarity compared to \( P_{\text{dual}} \) and \( P_{4.1} \). The \( P_{4.1} \) sensor shows lower similarity, as its resonance weakens with increased stretching, leading to higher statistical entropy in the spectrogram and, consequently, lower correlation. Comparing \( P_{3.1} \) and \( P_{\text{dual}} \), the \( P_{3.1} \) sensor experiences gradual changes in \( w_c \) across each segment, making it less distinguishable than the inherent \( w_c \) difference in the \( P_{\text{dual}} \) configuration.

\begin{table}[tbp!]
\centering
\caption{Mean Absolute Error (MAE) in mm for estimating \( \Delta L_I \), \( \Delta L_O \), and \( \Delta L_{\text{all}} \) using a gradient boosting regressor.}
\vspace{3pt}
\label{table:ML_result}
\begin{tabular}{c c c c}
\hline
       & \( \Delta L_I \)   & \( \Delta L_O \)   & \( \Delta L_{\text{all}} \) \\ \hline
\( P_{3.1} \) & 0.93 & 0.90 & 0.68 \\
\( P_{4.1} \)& 0.90 & 1.37 & 0.99 \\
\( P_{\text{dual}} \)   & 0.79 & 0.79 & 0.75 \\ \hline
\end{tabular}

\end{table}
\begin{figure}[tbp!]
    \centering 
            
    \begin{subfigure}[h]{0.32\linewidth}
        \centering
        \includegraphics[width=\linewidth]{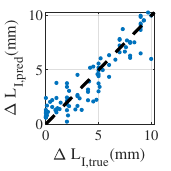}
        \caption{\( P_{3.1} \) - $L_I$}
    \end{subfigure}%
    \begin{subfigure}[h]{0.32\linewidth}
        \centering
        \includegraphics[width=\linewidth]{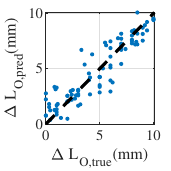}
        \caption{\( P_{3.1} \) - $L_O$}
    \end{subfigure}%
    \begin{subfigure}[h]{0.32\linewidth}
        \centering
        \includegraphics[width=\linewidth]{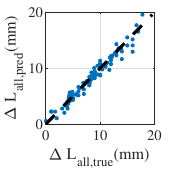}
        \caption{\( P_{3.1} \) - $L_{all}$}
    \end{subfigure}
    
    \begin{subfigure}[h]{0.32\linewidth}
        \centering
        \includegraphics[width=\linewidth]{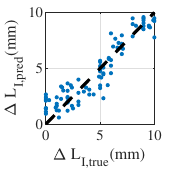}
        \caption{\( P_{4.1} \) - $L_I$}
    \end{subfigure}%
    \begin{subfigure}[h]{0.32\linewidth}
        \centering
        \includegraphics[width=\linewidth]{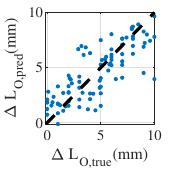}
        \caption{\( P_{4.1} \) - $L_O$}
    \end{subfigure}%
    \begin{subfigure}[h]{0.32\linewidth}
        \centering
        \includegraphics[width=\linewidth]{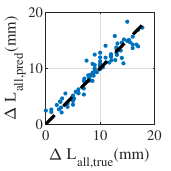}
        \caption{\( P_{4.1} \) - $L_{all}$}
    \end{subfigure}%
    
    \begin{subfigure}[h]{0.32\linewidth}
        \centering
        \includegraphics[width=\linewidth]{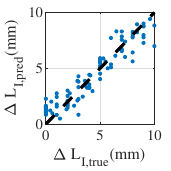}
        \caption{\( P_{\text{dual}} \) - $L_I$}
    \end{subfigure}%
    \begin{subfigure}[h]{0.32\linewidth}
        \centering
        \includegraphics[width=\linewidth]{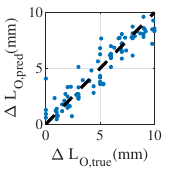}
        \caption{\( P_{\text{dual}} \) - $L_O$}
    \end{subfigure}%
    \begin{subfigure}[h]{0.32\linewidth}
        \centering
        \includegraphics[width=\linewidth]{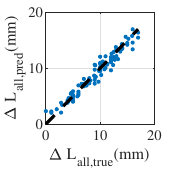}
        \caption{\( P_{\text{dual}} \) - $L_{all}$}
    \end{subfigure}%

    \vspace{5pt} 
    \caption{Predicted vs. actual stretch plot}
    \label{fig:MachinelearningPlot}
\end{figure}

\subsubsection{Machine Learning for Segment Stretch Differentiation}
\label{sec:ML}
To effectively leverage both \( F\text{-}U \) slope and resonance frequencies in machine learning, we extracted a feature vector containing the resonance peak frequency at each sampled flow speed, resulting in a 32-dimensional vector from each stretch configuration.

Using these feature data, we applied a gradient boosting regressor \cite{pedregosa2011scikit} to estimate the stretch of each segment and the total stretch. Out of 442 data points for each sensor, 80\% were used for training, and 20\% for testing. The test results, shown in \cref{fig:MachinelearningPlot} and \cref{table:ML_result}, illustrate the distribution of predicted versus true stretch values and their Mean Absolute Error (MAE). For total length estimation (\( L_{\text{all}} \)), the \( P_{3.1} \) sensor outperformed the others, while for segmented length estimation, \( P_{\text{dual}} \) performed better, as anticipated from \cref{sec:segmentedStretch}. Although the segmentation length error for \( P_{3.1} \) was slightly higher, it still achieved a MAE of less than 1 mm, making it an acceptable choice given its simpler, consistent corrugation pitch compared to the dual-pitch design. The \( P_{4.1} \) sensor exhibited the lowest performance, likely due to its weak resonance under stretch.



\subsection{Application for Finger Joint Angle Differentiation}

\begin{figure}[tbp!]
    \centering 
        
    \begin{subfigure}[h]{0.32\linewidth}
        \centering
        \includegraphics[width=0.9\linewidth]{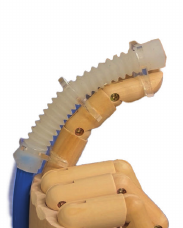}
        \caption{Overall bend}
    \end{subfigure}%
    \begin{subfigure}[h]{0.32\linewidth}
        \centering
        \includegraphics[width=0.9\linewidth]{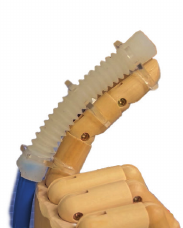}
        \caption{Inlet-side bend}
    \end{subfigure}%
    \begin{subfigure}[h]{0.32\linewidth}
        \centering
        \includegraphics[width=0.9\linewidth]{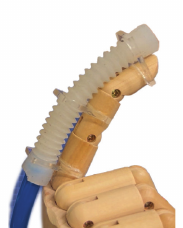}
        \caption{Outlet-side bend}
    \end{subfigure}

    \begin{subfigure}[h]{0.49\linewidth}
        \centering
        \includegraphics[width=0.9\linewidth]{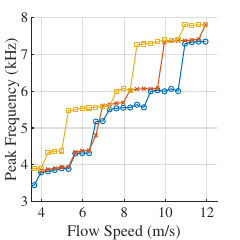}
        \caption{Peak Resonance Frequency}
    \end{subfigure}%
    \begin{subfigure}[h]{0.49\linewidth}
        \centering
        \includegraphics[width=0.9\linewidth]{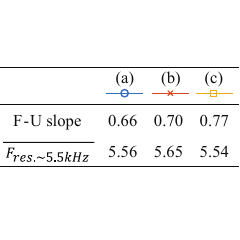}
    \end{subfigure}%

    \vspace{5pt} 
    \caption{Finger joint angle differentiation test with three different configurations (a-c).}
    \vspace{-10pt} 
    \label{fig:woodHand}
\end{figure}

Using the \( P_{3.1} \) sensor, we tested its ability to differentiate various articulated finger joint configurations. As shown in \cref{fig:woodHand}, resonance frequency alone may not be sufficient to distinguish cases such as both joints bending versus only the distal joint bending. The \( F\text{-}U \) slope, however, confirms findings from \cref{sec:segmentedStretch}, where no inlet-side stretch (\cref{fig:woodHand}c) exhibits the steepest \( F\text{-}U \) slope. We attempted to apply the machine learning model trained in \cref{sec:ML}, but its performance was limited, likely because it was trained on data from straight stretching rather than bending strain. A similar machine learning model incorporating bending of the sensors could be developed in future work to address this limitation.




\section{Conclusion}

We developed a soft corrugated tube sensor capable of estimating strain in each half segment. When air flows through the tube, the internal cavities induce vortex shedding, resulting in pressure oscillations that, when aligned with the tube’s standing wave resonance, produce a measurable acoustic tone. Since the vortex shedding frequency is linearly proportional to flow speed, the acoustic resonance frequency also increases linearly with flow speed. 

When the corrugated tube stretches, two factors affect the acoustic resonance: the resonance frequency decreases due to the standing wave effect, and the frequency-flow speed (\( F\text{-}U \)) slope shifts, as it is sensitive to changes in cavity width. Experiments show that the inlet-side cavity width has a dominant effect on the \( F\text{-}U \) slope, making it useful for estimating each half segment’s stretch. By sweeping the flow rate in a controlled manner, we captured resonance frequencies across flow speeds, enabling a machine learning algorithm (gradient boosting regressor) to estimate each half segment’s stretch. The dual-period tube design (3.1 mm and 4.18 mm corrugation periods) achieved the best results, with a mean absolute error (MAE) of 0.8 mm, while the single-period tube (3.1 mm period) provided satisfactory accuracy with an MAE of about 1 mm. Testing on a mannequin finger demonstrated the sensor’s ability to differentiate multiple two-joint configurations.

Currently, our proposed sensor has limitations. It may not perform well in environments with ambient resonant tones, though we consider such cases rare. Additionally, the sensor's noticeable acoustic tone makes it less suitable for environments with humans. However, it is well-suited for human-free settings, such as automated factories or outdoor fields. The current flowrate sweep takes 0.8 seconds, resulting in a maximum strain sampling rate of 1.25 Hz. Further studies could explore ways to optimize the sweep speed and flowrate range to improve the sampling rate. Finally, while an MAE of 1 mm may not meet the precision requirements for applications like surgical robotics, we believe it is acceptable for many other applications.


For future work, we aim to embed this sensor into a soft gripper to serve as a bending sensor, enabling accurate shape estimation even with external contacts. 
We anticipate that using a corrugated tube made from the same material will ensure seamless incorporation of the strain and bending sensor into the soft manipulator.
Additionally, we plan to explore corrugation designs that allow for larger strain sensing, accommodating significant bending in soft robotic applications. 

\bibliographystyle{IEEEtran}

\balance  

\bibliography{IEEEabrv,Biblio}   
\vspace{\baselineskip}


\end{document}